\documentclass[journal,11pt,onecolumn]{IEEEtran}
\usepackage{newtxtext,newtxmath}
\usepackage{hyperref}
\usepackage{subcaption}
\usepackage{supertabular}
\usepackage{longtable}
\usepackage{siunitx}
\usepackage{tabularx}
\usepackage{fleqn}
\usepackage{lipsum}
\usepackage{float}
\usepackage{xcolor}
\usepackage{enumitem}
\usepackage{hyperref}
\usepackage{graphicx}
\usepackage{amsmath}
\usepackage{geometry}

\ifCLASSINFOpdf
\else
\fi
\usepackage[blocks]{authblk}

\usepackage{setspace}
\title{Spacecraft Autonomous Decision-Planning for Collision Avoidance : a Reinforcement Learning Approach \\}

\author[1]{Nicolas Bourriez}
\author[1]{Adrien Loizeau}
\author[2,*]{Adam F. Abdin}

\affil[1]{CentraleSup\'{e}lec, Universit\'{e} Paris-Saclay, Gif-sur-Yvette, France.\vspace*{.2cm}}
\affil[2]{Laboratoire G\'{e}nie Industriel,  CentraleSup\'{e}lec, Universit\'{e} Paris-Saclay, Gif-sur-Yvette, France.\vspace*{.2cm}}

\affil[*]{Corresponding author. e-mail: adam.abdin@centralesupelec.fr}
\affil[+]{Speaker}

\begin{document}

\markboth{\fontsize{8}{5} \selectfont Preprint accepted in the 74$^{th}$ International Astronautical Congress (IAC) - Baku, Azerbaijan, 2-6 October 2023.}%
{Bourriez \MakeLowercase{\textit{et al.}}: Spacecraft Autonomous Decision-Planning for Collision Avoidance :\\ a Reinforcement Learning Approach }

\maketitle

\begin{abstract}
The space environment around the Earth is becoming increasingly populated by both active spacecraft and space debris. To avoid potential collision events, significant improvements in Space Situational Awareness (SSA) activities and Collision Avoidance (CA) technologies are allowing the tracking and maneuvering of spacecraft with increasing accuracy and reliability. However, these procedures still largely involve a high level of human intervention to make the necessary decisions. For an increasingly complex space environment, this decision-making strategy is not likely to be sustainable. Therefore, it is important to successfully introduce higher levels of automation for key Space Traffic Management (STM) processes to ensure the level of reliability needed for navigating a large number of spacecraft. These processes range from collision risk detection to the identification of the appropriate action to take and the execution of avoidance maneuvers. This work proposes an implementation of autonomous CA decision-making capabilities on spacecraft based on Reinforcement Learning (RL) techniques. A novel methodology based on a Partially Observable Markov Decision Process (POMDP) framework is developed to train the Artificial Intelligence (AI) system on board the spacecraft, considering epistemic and aleatory uncertainties. The proposed framework considers imperfect monitoring information about the status of the debris in orbit and allows the AI system to effectively learn stochastic policies to perform accurate Collision Avoidance Maneuvers (CAMs). The objective is to successfully delegate the decision-making process for autonomously implementing a CAM to the spacecraft without human intervention. This approach would allow for a faster response in the decision-making process and for highly decentralized operations.
\end{abstract}


%
\IEEEpeerreviewmaketitle

\section{\textbf{Introduction}}
\label{section:intro}

\IEEEPARstart{T}{he} current space environment around the Earth is becoming increasingly populated by both active spacecraft (satellites and launch vehicles) and space debris. Thousands of pieces of debris, measuring at least 10cm in diameter, and millions of pieces larger than 1cm, traveling at extremely high speeds, can significantly damage a spacecraft upon collision \cite{radtke2017interactions}. Collisions with space debris can generate more debris, which can then lead to further collisions, creating a chain reaction known as the Kessler syndrome \cite{krag20171}. This can lead to a severe threat to the long-term sustainability of Earth's orbits \cite{muelhaupt2019space}.

\par To minimize the risk of collisions with active or inactive objects in Earth's orbit, owners and operators (O/Os) of satellites must be aware of the collision risk to their assets \cite{le2018space} and need to implement proper actions to manage these risks. With mega-constellations rising and the significant increase in spacecraft orbiting the Earth, managing collision risks and developing proper strategies to improve space situational awareness (SSA) and space traffic management (STM) have become crucially important. Several challenges remain related to the accurate monitoring and tracking of space objects and the development of proper decision-support frameworks, leveraging advancements in Artificial Intelligence (AI) methods to use these data to make informed risk mitigation and operational management decisions.

As part of improving SSA capabilities, space agencies around the world are developing technologies for systems that can detect and track space objects and issue an alert when evasive action may be necessary. The global Space Surveillance Network (SSN) is an example of one such system used to track objects in Earth's orbit and monitor their trajectories \cite{horstmann2017investigation,mashiku2019recommended}. The SSN is a network of ground-based radar and optical sensors used to track objects in Earth's orbit. A physics simulator uses SSN observations to predict the evolution of the state of objects over time. Each satellite, also known as a target/protected object, is compared to all other objects in the catalog to detect a conjunction or a close approach. When a conjunction between the target and another object, usually referred to as the chaser/debris, is detected, the SSN propagated states become available, and a Conjunction Data Message (CDM) is produced, containing information about the event. This information includes the time of closest approach (TCA) and the probability of collision. At the issue of these warnings, the satellite's owners and operators have to decide whether to take action to avoid a collision with the available information one to two days before the TCA. To make this decision, they must assess the collision risk, including the probability of collision and the potential consequences of the collision, to plan and implement a collision avoidance maneuver (CAM). 

As the space environment becomes increasingly congested, the tasks of detecting, assessing, and planning maneuvers to avoid collisions become more challenging for manual human responses and existing decision processes. Traditionally, experts have been responsible for planning and executing spacecraft CAMs, a process that typically takes days to hours of preparation. However, due to the growing demands of STM, there is a need for intelligent onboard autonomous systems to handle spacecraft maneuvering tasks. These autonomous systems can manage tasks such as collision avoidance and station keeping on a larger scale with faster response times. Automating these processes can significantly enhance the safety and efficiency of space operations, ultimately contributing to a more sustainable and secure future in space operations \cite{hobbs2020taxonomy}. However, developing onboard autonomous collision avoidance systems is a challenging task. The optimal maneuver must balance multiple factors such as collision probability, propellant consumption, and mission objectives \cite{kim2012optimal}. Moreover, the optimal decisions may need to take into account the inherent uncertainties in objects' velocities, location, and monitoring information, increasing the problem's complexity \cite{denenberg2017debris}. In addition, the adoption of appropriate computational frameworks is crucial for efficiently handling large volumes of data and improving prediction and decision-making accuracy under uncertainty. Addressing these issues requires adequate AI-based decision-support models and algorithms capable of evaluating the trade-off between these different objectives and operational constraints and efficiently finding the optimal solutions.

Recent studies have started to propose optimization and learning algorithms for improving orbital collision risk predictions and developing collision avoidance planning and execution algorithms. Some researchers have used machine learning (ML) to improve the prediction of the probability of collision using historical conjunction data \cite{pinto2020towards, vasile2018artificial}, while others researchers have explored the use of deep learning to simulate the future states of objects and predict the probability of a collision \cite{sanchez2019ai,fernandez2021use}. Other works have proposed methodologies for automating the CAM execution using polynomial regression models \cite{ramaneti2021autonomous} or robust Bayesian framework  \cite{greco2021robust}. In addition, studies developed analytical \cite{gonzalo2021analytical, gonzalo2020collision} and semi-analytical \cite{gonzalo2019semi} models to calculate expressions for the orbit modification to implement autonomous CAM, as well as heuristic approaches such as Particle Swarm Optimisation (PSO) for on-board trajectory generation for STM \cite{lagona2022autonomous}.

While these methods provide a wide variety of tools to develop autonomous CAM frameworks, they often rely on physics-based models to derive adequate action. These models often require a significant number of simplifications compared to the actual environment dynamics, making the decision-making output dependent on the model quality and the assumptions made. More recently, other tools have been developed based on novel maneuver optimization algorithm that combines domain knowledge with Reinforcement Learning (RL) algorithms \cite{willis2016reinforcement}. The RL-based approach allows the model to explore the environment, learning to map states (e.g., collision risks) to optimal actions (e.g., CAMs) without requiring explicit models of the state transitions (physics-based models) and has recently shown promising results in spacecraft CAMs planning  \cite{gremyachikh2019space, qu2022spacecraft}. However, existing RL applications to spacecraft CAM are based on modeling the decision problem as a Markov Decision Process (MDP), which assumes that the agent is capable of having perfect access to the state of the environment (e.g., accurate debris position and velocity). In reality, however, this is not the case since monitoring data on the state of the environment is imperfect and is characterized by a range of uncertainties. Monitoring data on the status of satellites’ systems and orbital environments, such as the proximity to orbital debris and the position and velocity of the debris, are highly uncertain and do not necessarily represent the actual state of the system (i.e., the system state is not fully observable). Therefore, proper modeling of the CAM problem requires methods capable of dealing with partial observability of the status of the systems.

To this end, in this paper, we propose a novel RL-based approach for developing spacecraft autonomous CAM systems that consider aleatory and epistemic uncertainties and in which the decision-makers do not have complete knowledge of the status of the environment. We model the AI planning problem mathematically as a Partially Observable Markov Decision Process (POMDP) to take into account the problem's uncertainties and imperfect monitoring of the environment's status. Moreover, we propose a novel solution algorithm based on Deep Recurrent Q-Network (DRQN) capable of solving the proposed POMDP with continuous and infinite state space and discretized action space. To the best of our knowledge, this is the first implementation of the POMDP formalism to develop AI algorithms for spacecraft planning tasks.

The rest of the paper is organized as follows. Section~(\ref{section:intro}), presented the problem context, the state-of-the-art and the paper contributions. In Section~(\ref{section:model}) we describe the proposed RL planning model for autonomous CAM planning and execution. The model is mathematically described within a POMDP formalism capable of realistically representing the uncertainties in the orbital environment. Section~(\ref{section:solver}) details the DRQN solution algorithm used for training the AI agent. Section~(\ref{section:results}) presents the results and validation of the training model. Finally, Section~(\ref{section:conclusions}) discusses the conclusions and further research perspectives.

\section{\textbf{Reinforcement learning with partial observability for autonomous CAM planning and execution}}
\label{section:model}

We model the AI system with the function of planning and executing CAM decisions onboard the spacecraft (the agent) as a system that does not have explicit knowledge about the possible states of the objects in its surroundings (i.e., debris and non-cooperative spacecraft). Instead, The agent has only imperfect monitoring information from sensory monitoring data and external information sources. A novel RL approach is used to train the agent to autonomously make optimal CAM decisions that simultaneously minimize collision risk and fuel expenditure to perform the maneuvers. For this, we mathematically model the learning process as a Partially Observable Markov Decision Process (POMDP).

A POMDP is a generalization of Markov Decision Processes (MDPs) to decision-making situations in which the real system states are not fully observable (i.e., the monitoring data are insufficient to describe the system's real state). This realistic extension significantly increases the complexity of the model and requires advanced RL algorithms capable of finding the optimal solutions and training the agent in continuous and uncertain observation spaces. To the best of our knowledge, this is the first implementation of the POMDP formalism to develop AI algorithms for spacecraft planning tasks. In this section, we describe in detail the methodology developed in our approach for spacecraft autonomous CAM decision-planning.

\subsection{\textbf{Partially Observable Markov Decision Process (POMDP) collision avoidance maneuver model}}

The POMDP is a flexible mathematical framework for representing sequential decision problems \cite{POMDP_original_article, POMDP_second_article}. Unlike in MDPs, in POMDP, the autonomous AI agent cannot directly observe the state of the environment. Instead, the agent only has access to observations that are generated probabilistically based on previous actions and imperfect data acquired from sensory monitoring. The POMDP framework, thus, models the inherent uncertainty in the space collision avoidance problem to obtain optimal decisions under aleatory and epistemic uncertainties. \\

To model the spacecraft decision-learning process under a POMDP framework, in our work, the model is defined using the tuple ($\mathcal{S}$, $\mathcal{A}$, $\mathcal{T}$, $\mathcal{R}$, $\mathcal{O}$, $\mathcal{Z}$, $\mathcal{\gamma}$), where:

\begin{itemize}
    \item $\mathcal{S}$: represents the state-space of the environment. It is a continuous space that describes the spacecraft's and space debris' position and velocity vectors in three-dimensional space ($x$, $y$, $z$, $dx$, $dy$, $dz$). 
    \item $\mathcal{A}$: represents the action space of the spacecraft. In our model, the agent can only modify its position by applying impulsive thrust to change its velocity. The action space is discrete, with five possible thrust values, including 0.0, 0.01, 0.05, 0.1, and -0.05, in each of the three dimensions and at a specific time. As a result, the action space has a size of $5^4 = 625$. 
    \item $\mathcal{T}$: represents the environment's state transition probability. We consider that transitions within the model are unknown (i.e., no analytical functions exist to model the state transitions) and that the agent can only learn through interaction with the environment. 
    \item $\mathcal{R}$: represents the reward (or penalty) received by the agent after taking an action. In our framework, the reward received by the agent is based on three components: collision probability, fuel consumption, and trajectory deviation. Detailed explanations of these components are provided below.
    \item $\mathcal{O}$: represents the observation space of the environment. Similar to the state space, observations are continuous and are defined by the position and velocity vectors of the spacecraft and space debris ($x$, $y$, $z$, $dx$, $dy$, $dz$). 
    \item $\mathcal{Z}$: represents the observation model detailed below. The observation model is characterized by two types of uncertainties: epistemic uncertainties stemming from the SGP4 model and aleatory uncertainties resulting from the monitoring sensors. In our approach, both uncertainties are modeled as a Gaussian distribution. 
    \item $\mathcal{\gamma}$: represents the learning factor of the RL model ($\in [0, 1]$) and is set to 0.99 in our problem.\\
\end{itemize}

An essential component of the presented model is the observation model denoted $\mathcal{Z}$. $\mathcal{Z}$($o$ | $s$, $a$, $s_0$) is the probability or probability density of receiving observation $o$ about the debris characteristics in state $s_0$, given the previous state and action were $s$ and $a$, respectively. Information about the state may be inferred from the entire history of previous actions and observations and the initial information, $b_0$. Thus, in a POMDP, the agent's policy is a function mapping each possible history, $h_t$ = ($b_0$, $a_0$, $o_1$, $a_1$, $o_2$, $\ldots$, $a_{t-1}$, $o_t$), to an action. In some cases, each state's probability can be calculated based on the history of observations. This distribution is known as a belief ($b$), with $b_{t(s)}$ denoting the probability of state $s$. The belief is a sufficient statistic for optimal decision-making. \\

There exists a policy, $\pi$, such that when $a_t$ = $\pi$(bt), the expected cumulative reward is maximized for the POMDP. Given the POMDP model, each subsequent belief can be calculated using Bayes' rule. However, the exact update is computationally intensive, so approximate approaches such as particle filtering are usually used. For our model, the approach implemented to calculate the belief state is discussed in Section~(\ref{section:solver}). \\

A critical precursor to achieving the training objective is the formulation of an appropriate reward structure. The rewards serve as a critical feedback mechanism for guiding the learning process and shaping the agent's behavior. The cumulative sum of rewards obtained by the agent is a fundamental component of the loss function in our RL training model. This cumulative reward encapsulates the agent's performance to achieve optimal CAMs across each episode and serves as a critical signal for RL, facilitating the optimization of the agent's policy over time. In our formulation, we utilize a reward system characterized by predefined thresholds for various components, each of which contributes to the agent's cumulative sum reward. The key reward components and their associated thresholds are summarized in Table~\ref{tabreward}:

\begin{table}[h]
\centering
\caption{Reward Components and Threshold Values}
\begin{tabular}{lll}
\label{tabreward}
Reward Component         & Threshold Value  \\
\hline
Collision Probability    & $10^{-4}$ \\
Fuel Level               & 10 units \\
Trajectory Deviation (a) & 100 \\
Trajectory Deviation (e) & 0.01 \\
Trajectory Deviation (i) & 0.01 \\
Trajectory Deviation (W) & 0.01 \\
Trajectory Deviation (w) & 0.01 \\
\hline
\end{tabular}
\end{table}

\begin{itemize}
    \item \textit{Collision probability reward:} The agent receives negative rewards when the collision probability exceeds the threshold of $10^{-4}$, aligning with NASA's standards for collision risk mitigation in satellite operations.
\item \textit{Fuel level reward}: The fuel level is subject to a threshold of 10 units, imposing a negative reward for each action that consumes fuel; this ensures efficient fuel management.
\item \textit{Trajectory deviation reward}: Trajectory deviation is evaluated with respect to the differences in the first five osculating Keplerian elements (a, e, i, W, and w). Thresholds of 100, 0.01, 0.01, 0.01, and 0.01 are applied to each respective element, and the agent receives negative rewards proportional to its deviation from these desired values.

\end{itemize}

\subsection{\textbf{Collision avoidance data generation and simulation model}}

To ensure proper training of the spacecraft's autonomous decision-planning model, the quality of the data used is of significant importance. Particularly data related to conjunction events or near-conjunction scenarios to which the agent could be trained. Existing CDMs and Two Line Elements (TLEs) data sets can typically be used as training data. However, historical data on collision events are limited. Furthermore, collisions are typically catastrophic events that result in the destruction of the objects involved. This results in a lack of available data on the dynamics of a collision event. Therefore, available CDMs and TLEs data are not sufficient for training the RL agent to perform CAM optimally.

Instead, we rely on a custom simulator to generate simulated collision events with a wide range of customizable parameters and scenarios. This provides a much larger and more diverse dataset to train the agent, allowing it to generalize better and adapt to new situations. Additionally, the simulator allows for the collection of detailed data during the simulated collision events, providing a better understanding of the dynamics involved. 

\subsection{\textbf{Conjunction simulation model}}

We simulate conjunction events using an adaptation of the simulator developed in \cite{gremyachikh2019space}. This simulator uses Keplerian elements to instantiate debris objects and generate their positions within a specific range of the satellite. \\

\subsubsection{\textbf{Instantiating debris positions using Keplerian Elements}}

To generate collision scenarios between the protected spacecraft and space debris, the Pykep library \cite{dario_izzo_2017_1063506} was used to instantiate debris positions relative to the spacecraft. Pykep provides space flight mechanics computations based on perturbed Keplerian dynamics. The instantiation of the debris position can be defined using Keplerian elements such as semi-major axis, eccentricity, inclination, right ascension of the ascending node, argument of periapsis, and true anomaly. The model can, thus, generate debris positions within a specific range of the satellite, which is essential for creating different instances of trajectory scenarios. \\

\subsubsection{\textbf{Projection with SGP4 at each time step}}

To accurately simulate the motion of objects in space, we use the Simplified General Perturbations 4 (SGP4) model, which is widely used for orbit propagation. SGP4 takes into account the gravitational effects of the Earth, the Moon, and the Sun, as well as atmospheric drag and solar radiation pressure. Thus, the simulation model projects the debris trajectory using SGP4 at each time step. This allows us to accurately simulate the objects' positions and velocity and predict their future positions, including potential collisions between the objects and the satellite. \\

\subsubsection{\textbf{Retrograde collision reconstruction and debris velocity adjustment}}

In this section, we present a method for the retrograde reconstruction of collisions, starting from their conjunction points and backtracking to an initial time step. The methodology involves the instantiation of the environment, including the protected spacecraft's parameters and its osculating Keplerian elements. For each instantiated debris, the collision time is stochastically determined, followed by the projection of the protected object to the collision time using the SGP4 model. Subsequently, the debris's position is generated with a controlled proximity based on the projected object's position. Finally, the debris's velocity is adjusted in order to simulate the correct direction of motion leading to a collision based on a probability distribution.

\begin{enumerate}
\item \textit{Initialization of the environment}: At the start time, we instantiate the protected spacecraft, setting its characteristics such as radius, gravity ($\mu$), fuel level, and Earth's gravitational parameter ($\mu_{\text{central\_body}}$). Additionally, six osculating Keplerian elements are defined: semi-major axis ($a$) in meters; eccentricity ($e$) $ \in [0, 1)$; inclination ($i$) in radians; longitude of the ascending node ($W$) in radians; argument of periapses ($w$) in radians; and mean anomaly ($M$) in radians.

\item \textit{Debris instantiation}: for each debris that we choose to instantiate ($n_{\text{debris}}$),
   \begin{enumerate}
   \item The collision time is randomly determined between \textit{start\_time} and \textit{end\_time} following a uniform distribution.
   \item Once the collision time is established, we project our protected object up to that collision time using the SGP4 model. This generates a position vector and a velocity vector projected at that precise collision moment.
   \item The debris's position is instantiated based on the projected object's position, with a certain determined proximity. The method to ascertain this collision proximity is as follows: the generation of the debris's position follows a normal distribution centered around the projected object's position (expressed in meters), with a standard deviation $\sigma$ (a hyperparameter).
   \item Once the debris's position at collision instant $c$ is established, it is important to calculate the debris's velocity as well. However, to update the velocity vector of debris in space, we simulate the correct direction of motion for our debris, which will collide with the protected object according to a specific probability distribution.\\
   \end{enumerate}
\end{enumerate}

We generate the debris's velocity following Eq.~(\ref{eq1}):
\begin{align}
\label{eq1}
& rotate\_velocity(\text{vel}, \text{pos}, \theta) = \nonumber \\ 
&\|\text{vel}\| \cdot \left( \frac{\cos(\theta)}{\|\text{vel}\|} \cdot \text{vel} + \frac{\sin(\cos(\theta)}{\|w\|} \cdot w \right)
\end{align}
where:
\begin{align*}
vel &= \text{velocity vector} \\
pos &= \text{position vector} \\
\theta &= \text{random angle in radians} \\
w &= pos \times vel
\end{align*}
It is important to point out that $\theta$ is randomly chosen between two possible ranges of angles, each chosen with equal probability $\frac{1}{2}$. These ranges are hyperparameters, acting as the minimum angle between debris and protected object at collision time (radians) ($\leq \pi/4$ by default). Finally, some noise is added to the xdebris' velocity in order to simulate variations or uncertainties in the magnitude of the velocity, which can be due to factors such as measurement errors, external influences, or modeling inaccuracies. This is expressed in the following equation:

$$\mathbf{v}_{\text{new}} = \mathbf{v} \times \mathcal{N}(1, \sigma_{\text{vr}})$$

where:
\begin{itemize}
    \item $\mathbf{v}_{\text{new}}$ is the new velocity vector after applying the scaling.
    \item $\mathbf{v}$ is the original velocity vector.
    \item $\mathcal{N}(1, \sigma_{\text{vr}})$ is a normal distribution with mean 1 and standard deviation $\sigma_{\text{vr}}$.
    \item $\sigma_{\text{vr}}$ is the standard deviation controlling the randomness of the scaling.\\
\end{itemize}

Finally, in order to comprehensively assess the performance and robustness of the algorithms developed for collision prediction and avoidance in dynamic environments, a large number of scenarios are generated through a systematic approach, where the parameters governing the behaviors of debris and protective systems are drawn from probability distributions. The developed approach involves a controlled variation of environmental factors that can influence the complexity of collision scenarios. To effectively simulate a range of plausible conditions, key parameters such as the number of debris particles, the temporal span of the environment, the debris positional deviations, the velocity ratio fluctuations, and the minimum angle between debris and protective systems at the point of potential collision, are stochastically determined.

\section{\textbf{Solution algorithm}}
\label{section:solver}

\begin{figure*}[h]
\centering
\includegraphics[scale = 0.42]{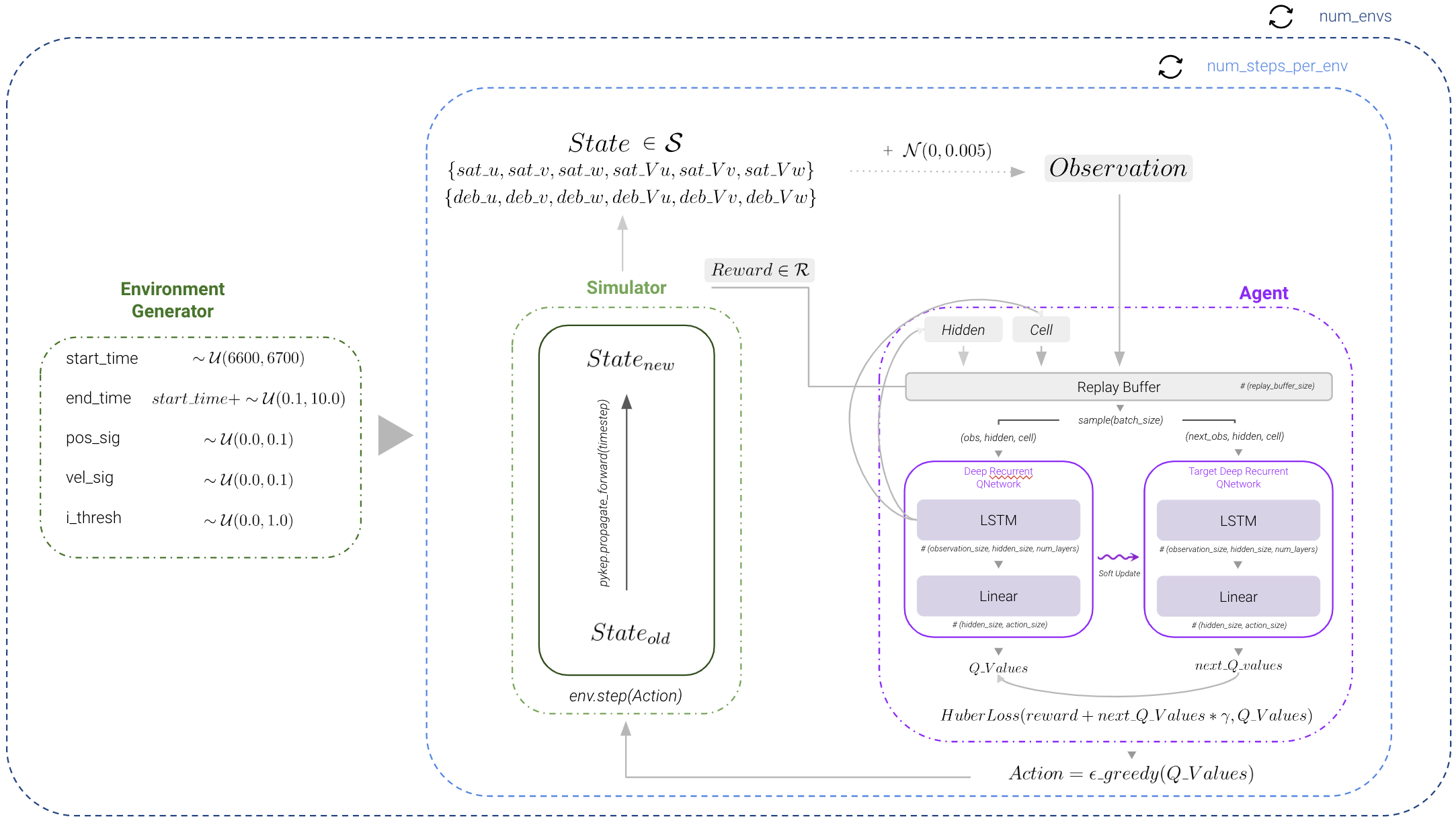}
\caption{Architecture of the proposed Deep Recurrent Q-Network (DRQN) algorithm}
\label{figDRQN}
\end{figure*}

To solve our proposed POMDP-CAM planning and execution model, an appropriate solver should be implemented. Different RL algorithms can be developed to train the AI system proposed. Given the problem-specific characteristics, Deep Recurrent Q-Network (DRQN) has been found to be the most capable algorithm for training the autonomous CAM system. This section details the DRQN algorithm and discusses its advantages and limitations for training the agent in the context of our problem.

\subsection{\textbf{Deep Recurrent Q-Network (DRQN)}}
The proposed CAM learning model is a continuous model-free POMDP with infinite state space and discretized action space. Only a few state-of-the-art algorithms exist to solve these types of models with infinite $mathcal{S}$ calls and a non-linear function approximator. Deep Q-Network (DQN) has shown promising results for solving the more simple MDPs with similar modeling characteristics. Recent research works have proposed extensions for DQN for solving POMDP problems by adding a Recurrent Neural Network (RNN) layer to the Q-Network \cite{hausknecht2015deep}.  The DRQN can serve as a non-linear function approximator for our proposed POMDP-CAM planning model since the state space of space debris is extremely large, and traditional linear approximators cannot capture the complex relationships between different state variables. DRQNs, on the other hand, can learn and represent these non-linear relationships effectively. \\

Furthermore, the RNN layer in the DRQN architecture plays the role of belief updater by keeping track of the history of past observations and actions. Thus, the latent, hidden state can be considered equivalent to a classical ``belief'' in the POMDP model. For instance, if the agent observes a sudden change in the position of a piece of debris, the RNN layer can consider past observations and update the agent's belief on whether this change will likely result in a collision.

\subsubsection{\textbf{Solver Architecture}}

For each conditional environment, a set of stochastic parameters is generated by the environment generator in order to simulate the debris and the protected object paths from start time to end time, with a certain probability of collision. For each step in this environment, the simulator updates its state $s$. Gaussian noise is added to simulate the model's imperfect monitoring (partially observability), translating it to the observation $o$. This observation is passed to the agent, i.e., the DRQN algorithm, which will take action accordingly. 
 
The implementation of the DRQN consists of two main components \cite{hausknecht2015deep}: an RNN layer (in that case, a Long Short-Term Networks (LSTM)) and a fully connected layer that approximates the Q-values for each action, given the current state as seen in Figure~(\ref{figDRQN}). The LSTM layer helps the network maintain a belief of past observations and actions. Furthermore, a ReplayBuffer and a Target Q-Network are added to enhance the learning process and address some of the DRQN's limitations. On one side, the ReplayBuffer is a memory buffer that stores past experiences (i.e., observation, action, reward, following observation, terminal) and randomly samples a batch of experiences to train the DRQN. This technique helps de-correlate the data from identical sequences and avoid overfitting when the same data is used multiple times. By storing past experiences, the ReplayBuffer also allows the DRQN to learn from experiences that occurred earlier in training, thus increasing the efficiency of the learning process. On the other side, Target Q-Network is a separate network that is softly updated with the weights of the DRQN \cite{lillicrap2015continuous}. This network aims to provide a more stable and consistent target for the DRQN to learn from. Without the Target Q Network, the DRQN would be learning from a constantly changing estimate. By soft updating the Target Q Network, the DRQN can learn from a more stable and consistent target, which helps accelerate the learning process and improve the final performance. \\

\subsubsection{\textbf{Loss Function}}
The selection of an appropriate loss function is essential in training deep RL models, particularly in the presence of outliers. For the proposed DRQN implementation, we employ the Huber loss function on the cumulative sum of rewards. This choice is driven by the necessity to robustly handle outliers, which can significantly impact our learning process. Huber loss, also known as smooth mean absolute error, offers the advantage of being less sensitive to extreme reward values, a common occurrence in complex environments featuring rare and substantial rewards or penalties. It is a combination of mean squared error (MSE) and mean absolute error (MAE), which allows it to be robust to outliers while maintaining smoothness and differentiability. Mathematically, the Huber loss function is defined as:

$$ L_{\delta}(a) = \begin{cases}
\frac{1}{2} a^2 & \text{for } |a| \leq \delta, \\
\delta (|a| - \frac{1}{2}\delta) & \text{otherwise.}
\end{cases}
$$

Where:
\begin{itemize}
 \item \(a\) is the difference between the predicted and actual values, \(a = y_{\text{pred}} - y_{\text{true}}\).
 \item \(\delta\) is a hyperparameter that determines the threshold at which the loss function changes from quadratic to linear. It should be chosen carefully as it affects the model's performance.
\end{itemize}

The Huber loss function behaves like MSE when the difference \(|a|\) is smaller than a threshold \(\delta\), and MAE when the difference \(|a|\) is greater than \(\delta\). This allows it to handle outliers more robustly than MSE, which squares the differences and thus gives more weight to larger differences. Additionally, the Huber loss function is differentiable at 0, which makes it suitable for optimization algorithms that require smooth and differentiable functions, such as gradient descent. \\

\subsubsection{\textbf{Constraints}}
One of the main challenges of using a DRQN is the algorithm's sensitivity to hyperparameters, particularly the $\texttt{tau}$ and $\texttt{replay\_buffer\_size}$ values. Small changes in the hyperparameter values could cause the network to diverge, making it difficult to train. Therefore, we perform a thorough investigation using hyperparameter sweeps to find the best values of these parameters. The results of hyperparameter tuning are discussed in the next section.

\section{\textbf{Results and discussion}} 
\label{section:results}

This section presents the initial results of the study. We first focus on presenting the results related to the efficacy of the DRQN model in learning optimal CAM policies. Given its inherent complexity, a central focus of this evaluation section is to elucidate the strategy for attaining solution convergence, both in a single training environment and across a large number of environments. The aim is to demonstrate the adaptability and robustness of our DRQN model to learn optimal CAM planning and execution policies under a variety of circumstances.

\subsection{\textbf{Hyperparameters tuning}}

\begin{figure}[htp!]
\begin{minipage}[b]{0.5\linewidth}
  \centering
  \subfloat[Grid Search over 1 environment]{
    \includegraphics[scale=0.1]{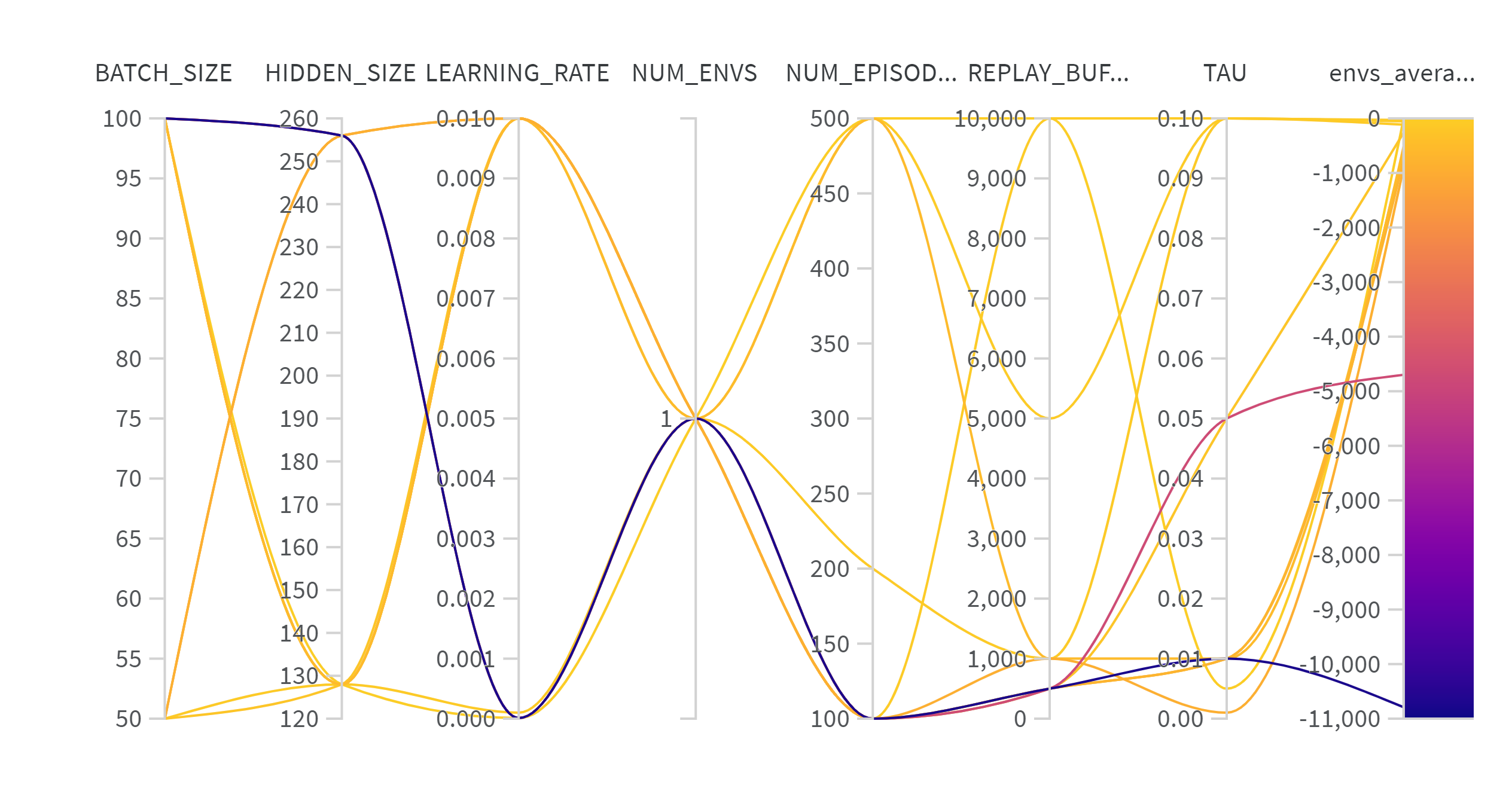}
    \label{fig:grid_1}
  }
\end{minipage}
\begin{minipage}[b]{0.5\linewidth}
  \centering
  \subfloat[Grid Search over 200 environments]{
    \includegraphics[scale=0.1]{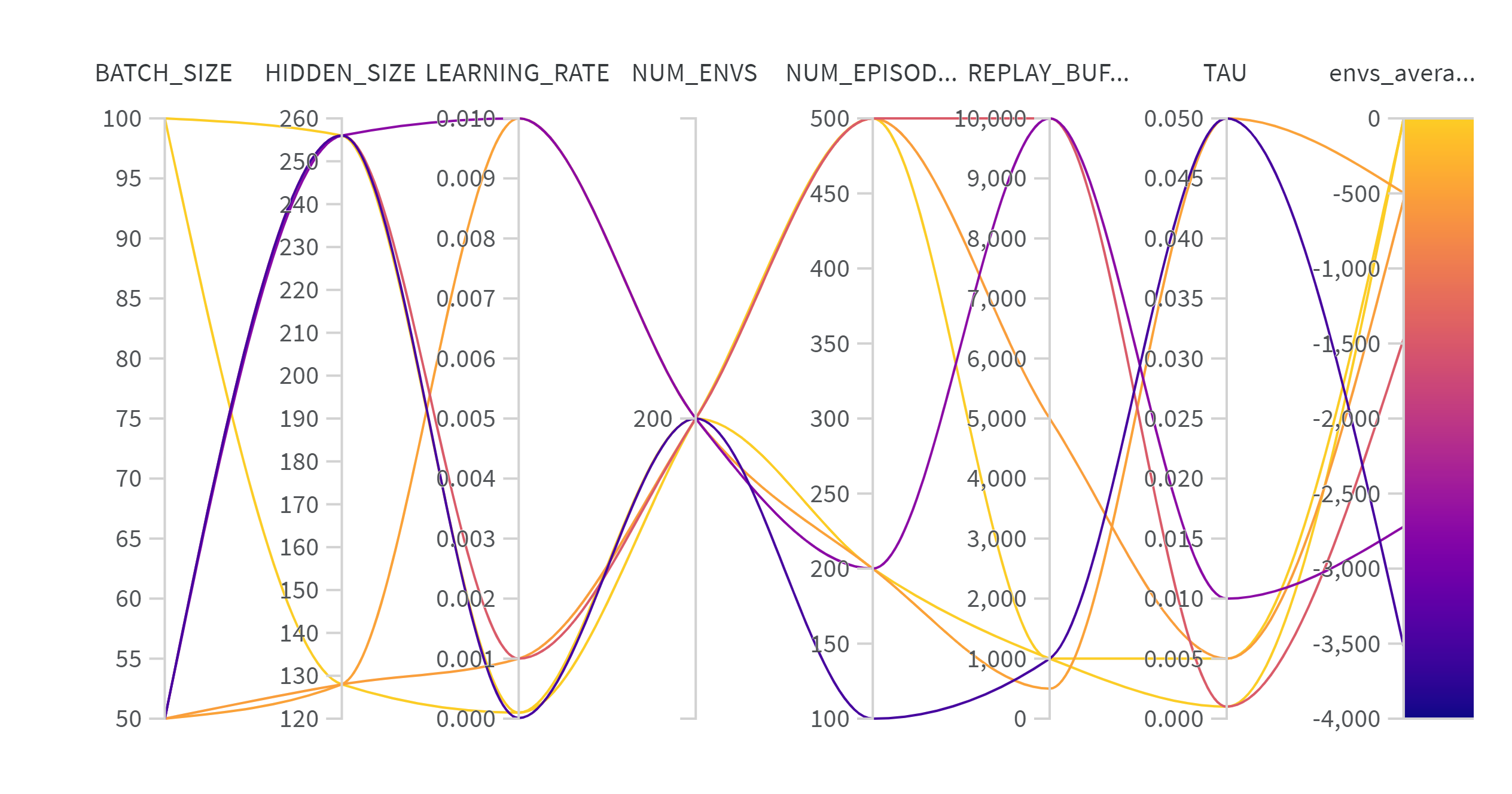}
    \label{fig:grid_200}
  }
\end{minipage}
\caption{Model hyperparameter tuning.}
\label{modelvalidation}
\end{figure}

We employ a grid search method for hyperparameter tuning of the model. Grid search operates by exhaustively evaluating a predefined set of hyperparameter values over a structured grid. Each unique combination of hyperparameters is systematically assessed, allowing for a comprehensive exploration of the hyperparameter space. In our problem, grid search enabled us to methodically examine various hyperparameter configurations, including neural network (NN) architecture specifications (e.g., hidden size layer), training parameters (e.g., learning rate, batch size), and exploration strategies (e.g., epsilon-greedy exploration). By systematically evaluating these hyperparameter choices on the average reward sum, we sought to identify the optimal configuration that would yield improved convergence and learning performance for our DRQN model. Figures (\ref{fig:grid_1}) and (\ref{fig:grid_200}) illustrate the grid search made for both 1 and 200 environments.



\subsection{\textbf{Training of Spacecraft CA agent}}
The process of hyperparameter optimization allowed the identification of the best parameter configurations. Subsequently, the next step is to select a configuration that would consistently converge across different stages of training. The hyperparameters selected for both training scenarios (under one environment and under 200 training environments) can be found in Tables (\ref{tab:hyperparameters-1-env}) and (\ref{tab:hyperparameters-200-env}), respectively. \\
\begin{table}[htp!]
\begin{minipage}[b]{0.5\linewidth}
\centering
\captionsetup{justification=centering}
\caption{Hyperparameters used for evaluation \\ on 1 environment}
\label{tab:hyperparameters-1-env}
\begin{tabular}{ll}
\hline
Hyperparameter           & Value          \\
\hline
Batch size             & 50       \\
Hidden size            & 128      \\
Learning rate          & 0.0001   \\
Number of episodes     & 200      \\
Replay buffer size     & 1,000    \\
Tau                    & 0.1      \\
Number of environments & 1        \\
\hline
\end{tabular}
\end{minipage}%
\begin{minipage}[b]{0.5\linewidth}
\centering
\captionsetup{justification=centering}
\caption{Hyperparameters used for evaluation \\ on 200 environments}
\label{tab:hyperparameters-200-env}
\begin{tabular}{ll}
\hline
Hyperparameter           & Value          \\
\hline
Batch size             & 100           \\
Hidden size            & 128          \\
Learning rate          & 0.0001       \\
Number of environments & 1            \\
Number of episodes     & 200          \\
Replay buffer size     & 1,000        \\
Tau                    & 0.1          \\
\hline
\end{tabular}
\end{minipage}
\end{table}

\begin{figure}[htp!]
\begin{minipage}[b]{0.5\linewidth}
  \centering
  \subfloat[Results of training on one environment]{
    \includegraphics[scale=0.045]{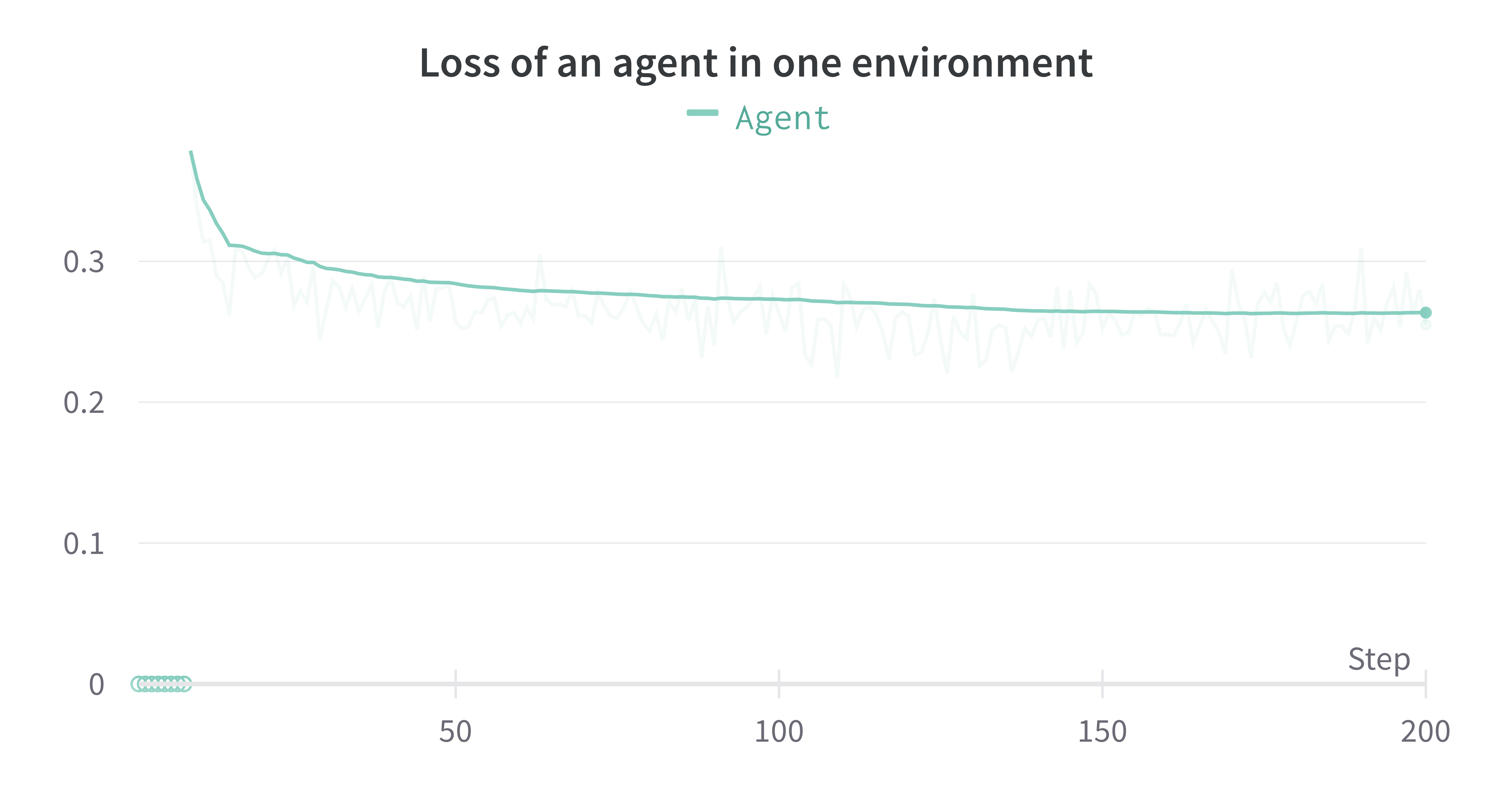}
    \label{fig:one_env}
  }
\end{minipage}
\begin{minipage}[b]{0.5\linewidth}
  \centering
  \subfloat[Results of training on 200 environments]{
    \includegraphics[scale=0.045]{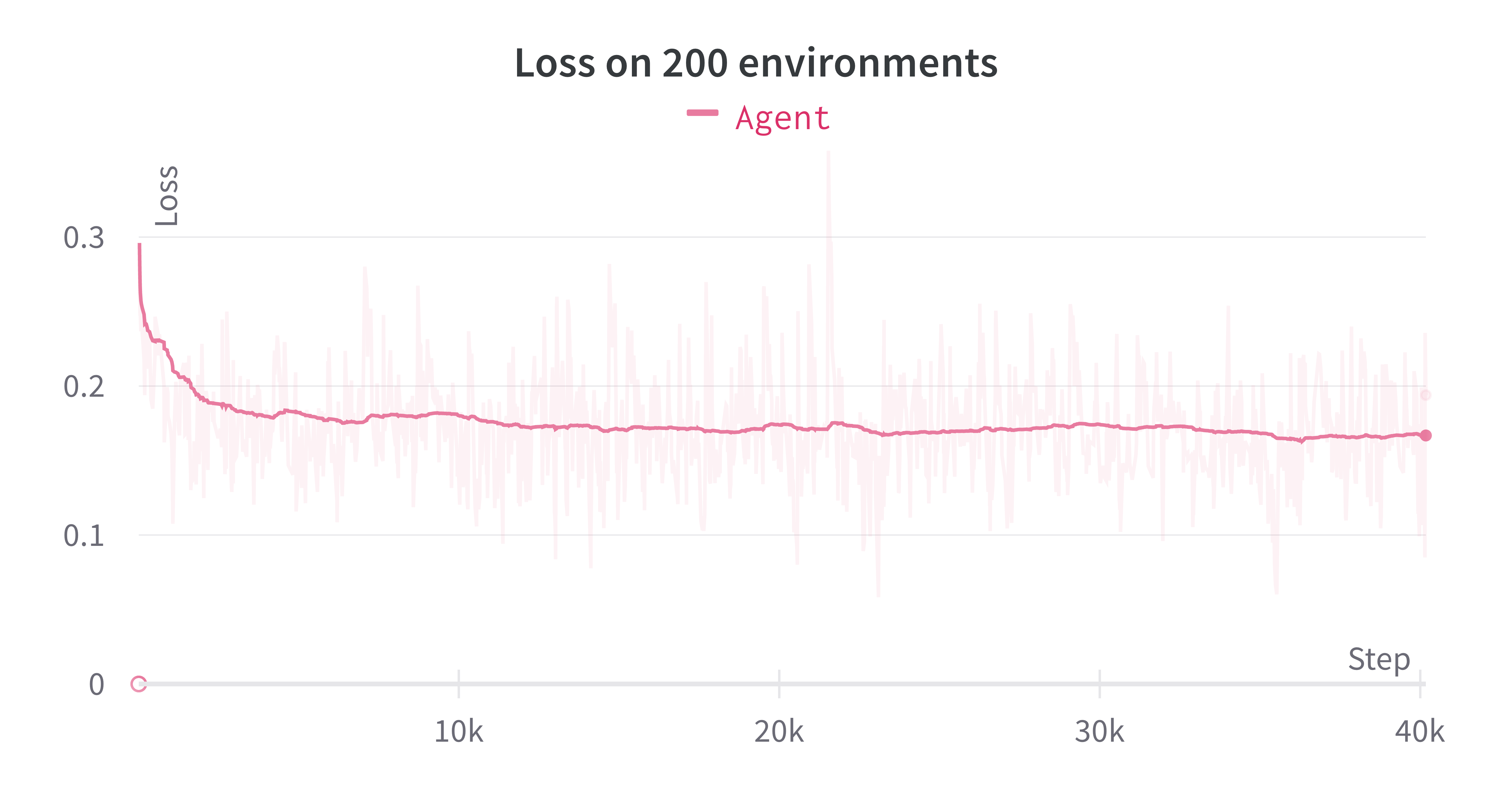}
    \label{fig:200_env}
  }
\end{minipage}
\caption{Loss of an agent (spacecraft autonomous CAM system) trained on different environments: decreasing loss signifies the agent's improvement in learning to perform optimal CAM planning and executions.}
\label{modelvalidation2}
\end{figure}

To evaluate the performance of the DRQN agent, we compare it to a baseline approach that uses a simple threshold to trigger a collision avoidance maneuver. Figure \ref{fig:one_env} illustrates the agent's loss trajectory over each training step within a single collision avoidance environment. These results show that the agent is successfully learning to autonomously conduct CAM maneuvers, as it illustrates the progressive improvement in maximizing its cumulative reward over the simulations. Furthermore, the training is extended to encompass a more complex scenario involving 200 distinct environments. The results show that the agent trained with this configuration attained better training results than those under a single training environment. Figure \ref{fig:200_env} shows the loss trajectory and the average cumulative reward profile of one of the successful agents in this more complex multi-environment setting. These results show that our proposed approach is capable of training an autonomous CAM system to plan and execute CAMs effectively, as evident by the significantly decreasing loss funtion. The trained spacecraft AI system performed well in terms of collision avoidance effectiveness, fuel consumption, and computational efficiency. In particular, the spacecraft was able to adapt to changing environments and make more efficient maneuvers in response to the available information. It is worth noting that the prospect of fine-tuning hyperparameters in future work remains an avenue for further enhancing the capabilities of our DRQN algorithm.



\section{\textbf{Conclusion}s}
\label{section:conclusions}

While significant improvements in Space Situational Awareness (SSA) activities and Collision Avoidance (CA) technologies are allowing for tracking and maneuvering spacecraft away from potential debris collision risks with increasing accuracy and reliability, these procedures still primarily involve a high level of human intervention to make the necessary decisions. This decision-making strategy will not be sustainable for an increasingly complex space environment.  It is, therefore, important to successfully introduce higher levels of automation for key Space Traffic Management (STM) processes to ensure the reliability needed for navigating large constellations of spacecraft. These processes include collision risk detection, the identification of the appropriate action to take, and the execution of avoidance maneuvers. 
In this work, we developed an implementation of autonomous CA capabilities for spacecraft based on Reinforcement Learning (RL) techniques. We propose to model the spacecraft CA training model using a novel Partially Observable Markov Decision Process (POMDP) solved with an efficient Deep Recurrent Q-Network (DRQN) algorithm. This allows the proper training of the AI system onboard the spacecraft as a system with imperfect monitoring information on all the possible states of the objects in its surroundings. This is particularly relevant to the practical setting of SSA for managing collision risk considering uncertainties in space debris positions and velocities.

The model proposed takes as input the current state of the environment, which includes the positions and velocities of the satellite and debris, as well as the CDM information, and outputs an action corresponding to a maneuver to avoid a potential collision. The agent is trained to maximize the expected cumulative reward over time. The reward function used includes the distance between the satellite and the debris, the probability of collision, and the fuel cost of the maneuver. Our results demonstrate the potential of using deep RL methods, such as DRQN, for developing autonomous CA systems in space. This approach could help mitigate the increasing risk of collisions with space debris and ensure the safety of space assets.  To the best of our knowledge, this is the first implementation of the POMDP formalism and the DRQN solution algorithm to develop AI systems for spacecraft planning tasks. 

Finally, several factors should be considered for the applicability and robustness of our approach in practical implementations. First, it should be noted that our model was trained on synthetic data generated to approximate realistic assumptions. However, the actual performance of our model in operational settings may deviate due to variations in the physical characteristics of space debris, satellite trajectories, and other environmental factors. Consequently, care should be taken to validate the proposed approach in controlled settings before practical implementations. Second, the precision and accuracy of CA estimations hinge on exogenous variables, such as the reliability of satellite position and velocity predictions delivered by orbital propagators like the Simplified General Perturbations 4 (SGP4) model. Any errors in these models can influence the performance of the collision avoidance system. Thus, continuous efforts to enhance the precision of these systems remain pivotal to enhancing the efficacy of our approach.

\section*{\textbf{Authors contributions}}

A.Abdin conceived the original idea of the study, the technical methodology proposed, and supervised the work. A.Loizeau and N.Bourriez contributed equally to developing and implementing the software, writing the first draft of the paper, and providing the initial analysis of the results. A.Abdin adapted and revised the manuscript.

\section*{\textbf{Acknowledgements}}
The authors would like to thank Matthieu Roux (Ph.D. candidate at the Laboratory of Industrial Engineering, CentraleSup\'{e}lec) for his valuable insights and suggestions on the implementation of the methodology proposed.

\ifCLASSOPTIONcaptionsoff
  \newpage
\fi

\newpage

\bibliographystyle{IEEEtran}
\bibliography{IEEEexample}

\begin{thebibliography}{10}
\providecommand{\url}[1]{#1}
\csname url@samestyle\endcsname
\providecommand{\newblock}{\relax}
\providecommand{\bibinfo}[2]{#2}
\providecommand{\BIBentrySTDinterwordspacing}{\spaceskip=0pt\relax}
\providecommand{\BIBentryALTinterwordstretchfactor}{4}
\providecommand{\BIBentryALTinterwordspacing}{\spaceskip=\fontdimen2\font plus
\BIBentryALTinterwordstretchfactor\fontdimen3\font minus \fontdimen4\font\relax}
\providecommand{\BIBforeignlanguage}[2]{{%
\expandafter\ifx\csname l@#1\endcsname\relax
\typeout{** WARNING: IEEEtran.bst: No hyphenation pattern has been}%
\typeout{** loaded for the language `#1'. Using the pattern for}%
\typeout{** the default language instead.}%
\else
\language=\csname l@#1\endcsname
\fi
#2}}
\providecommand{\BIBdecl}{\relax}
\BIBdecl

\bibitem{radtke2017interactions}
J.~Radtke, C.~Kebschull, and E.~Stoll, ``Interactions of the space debris environment with mega constellations—using the example of the oneweb constellation,'' \emph{Acta Astronautica}, vol. 131, pp. 55--68, 2017.

\bibitem{krag20171}
H.~Krag, M.~Serrano, V.~Braun, P.~Kuchynka, M.~Catania, J.~Siminski, M.~Schimmerohn, X.~Marc, D.~Kuijper, I.~Shurmer \emph{et~al.}, ``A 1 cm space debris impact onto the sentinel-1a solar array,'' \emph{Acta Astronautica}, vol. 137, pp. 434--443, 2017.

\bibitem{muelhaupt2019space}
T.~J. Muelhaupt, M.~E. Sorge, J.~Morin, and R.~S. Wilson, ``Space traffic management in the new space era,'' \emph{Journal of Space Safety Engineering}, vol.~6, no.~2, pp. 80--87, 2019.

\bibitem{le2018space}
S.~Le~May, S.~Gehly, B.~Carter, and S.~Flegel, ``Space debris collision probability analysis for proposed global broadband constellations,'' \emph{Acta Astronautica}, vol. 151, pp. 445--455, 2018.

\bibitem{horstmann2017investigation}
A.~Horstmann and E.~Stoll, ``Investigation of propagation accuracy effects within the modeling of space debris,'' in \emph{7th European Conference on Space Debris}, 2017.

\bibitem{mashiku2019recommended}
A.~K. Mashiku and M.~D. Hejduk, ``Recommended methods for setting mission conjunction analysis hard body radii,'' in \emph{2019 AAS/AIAA Astrodynamics Specialist Conference}, no. GSFC-E-DAA-TN71115-1, 2019.

\bibitem{hobbs2020taxonomy}
K.~L. Hobbs and E.~M. Feron, ``A taxonomy for aerospace collision avoidance with implications for automation in space traffic management,'' in \emph{AIAA Scitech 2020 Forum}, 2020, p. 0877.

\bibitem{kim2012optimal}
E.-H. Kim, H.-D. Kim, and H.-J. Kim, ``Optimal solution of collision avoidance maneuver with multiple space debris,'' \emph{Journal of Space Operations}, vol.~9, no.~3, pp. 20--31, 2012.

\bibitem{denenberg2017debris}
E.~Denenberg and P.~Gurfil, ``Debris avoidance maneuvers for spacecraft in a cluster,'' \emph{Journal of Guidance, Control, and Dynamics}, vol.~40, no.~6, pp. 1428--1440, 2017.

\bibitem{pinto2020towards}
F.~Pinto, G.~Acciarini, S.~Metz, S.~Boufelja, S.~Kaczmarek, K.~Merz, J.~A. Martinez-Heras, F.~Letizia, C.~Bridges, and A.~G. Baydin, ``Towards automated satellite conjunction management with bayesian deep learning,'' \emph{arXiv preprint arXiv:2012.12450}, 2020.

\bibitem{vasile2018artificial}
M.~Vasile, V.~Rodr{\'\i}guez-Fern{\'a}ndez, R.~Serra, D.~Camacho, and A.~Riccardi, ``Artificial intelligence in support to space traffic management,'' in \emph{68th International Astronautical Congress}, 2017.

\bibitem{sanchez2019ai}
L.~S{\'a}nchez, M.~Vasile, and E.~Minisci, ``Ai to support decision making in collision risk assessment,'' in \emph{70th International Astronautical Congress}, 2019.

\bibitem{fernandez2021use}
L.~S. Fernandez-Mellado and M.~Vasile, ``On the use of machine learning and evidence theory to improve collision risk management,'' \emph{Acta Astronautica}, vol. 181, pp. 694--706, 2021.

\bibitem{ramaneti2021autonomous}
K.~A. Ramaneti, C.~Krishna, A.~Ahmed, and R.~Rajesh, ``Autonomous space debris collision avoidance system,'' in \emph{Advances in Automation, Signal Processing, Instrumentation, and Control: Select Proceedings of i-CASIC 2020}.\hskip 1em plus 0.5em minus 0.4em\relax Springer, 2021, pp. 2489--2501.

\bibitem{greco2021robust}
C.~Greco, L.~S{\'a}nchez, M.~Manzi, and M.~Vasile, ``A robust bayesian agent for optimal collision avoidance manoeuvre planning,'' in \emph{8th European Conference on Space Debris}, 2021.

\bibitem{gonzalo2021analytical}
J.~L. Gonzalo, C.~Colombo, and P.~Di~Lizia, ``Analytical framework for space debris collision avoidance maneuver design,'' \emph{Journal of Guidance, Control, and Dynamics}, vol.~44, no.~3, pp. 469--487, 2021.

\bibitem{gonzalo2020collision}
J.~Gonzalo~G{\`o}mez, C.~Colombo \emph{et~al.}, ``Collision avoidance algorithms for space traffic management applications,'' in \emph{71st International Astronautical Congress}, 2020.

\bibitem{gonzalo2019semi}
J.~Gonzalo~Gomez, C.~Colombo, P.~Di~Lizia \emph{et~al.}, ``A semi-analytical approach to low-thrust collision avoidance manoeuvre design,'' in \emph{70th International Astronautical Congress}, 2019.

\bibitem{lagona2022autonomous}
E.~Lagona, S.~Hilton, A.~Afful, A.~Gardi, and R.~Sabatini, ``Autonomous trajectory optimisation for intelligent satellite systems and space traffic management,'' \emph{Acta Astronautica}, vol. 194, pp. 185--201, 2022.

\bibitem{willis2016reinforcement}
S.~Willis, D.~Izzo, and D.~Hennes, ``Reinforcement learning for spacecraft maneuvering near small bodies,'' in \emph{AAS/AIAA Space Flight Mechanics Meeting}, vol. 158, 2016, pp. 1351--1368.

\bibitem{gremyachikh2019space}
L.~Gremyachikh, D.~Dubov, N.~Kazeev, A.~Kulibaba, A.~Skuratov, A.~Tereshkin, A.~Ustyuzhanin, L.~Shiryaeva, and S.~Shishkin, ``Space navigator: A tool for the optimization of collision avoidance maneuvers,'' \emph{arXiv preprint arXiv:1902.02095}, 2019.

\bibitem{qu2022spacecraft}
Q.~Qu, K.~Liu, W.~Wang, and J.~L{\"u}, ``Spacecraft proximity maneuvering and rendezvous with collision avoidance based on reinforcement learning,'' \emph{IEEE Transactions on Aerospace and Electronic Systems}, vol.~58, no.~6, pp. 5823--5834, 2022.

\bibitem{POMDP_original_article}
L.~P. Kaelbling, M.~L. Littman, and A.~R. Cassandra, ``Partially observable markov decision processes for artificial intelligence,'' in \emph{KI-95: Advances in Artificial Intelligence}, I.~Wachsmuth, C.-R. Rollinger, and W.~Brauer, Eds.\hskip 1em plus 0.5em minus 0.4em\relax Berlin, Heidelberg: Springer Berlin Heidelberg, 1995, pp. 1--17.

\bibitem{POMDP_second_article}
S.~Thrun, W.~Burgard, and D.~Fox, \emph{Probabilistic Robotics (Intelligent Robotics and Autonomous Agents)}, 01 2005.

\bibitem{dario_izzo_2017_1063506}
\BIBentryALTinterwordspacing
D.~Izzo, ``esa/pykep: Major update.'' November 2017, https://doi.org/10.5281/zenodo.1063506 [Accessed: January-2023. [Online]. Available: \url{https://doi.org/10.5281/zenodo.1063506}
\BIBentrySTDinterwordspacing

\bibitem{hausknecht2015deep}
M.~Hausknecht and P.~Stone, ``Deep recurrent q-learning for partially observable mdps,'' in \emph{2015 aaai fall symposium series}, 2015.

\bibitem{lillicrap2015continuous}
T.~P. Lillicrap, J.~J. Hunt, A.~Pritzel, N.~Heess, T.~Erez, Y.~Tassa, D.~Silver, and D.~Wierstra, ``Continuous control with deep reinforcement learning,'' \emph{arXiv preprint arXiv:1509.02971}, 2015.

\end{thebibliography}

\end{document}